\documentclass[11pt]{article}
\usepackage[margin=1in]{geometry}
\usepackage{amsmath,amssymb,amsfonts}
\usepackage{graphicx}
\usepackage{booktabs}
\usepackage{tikz}
\usetikzlibrary{arrows.meta,positioning,patterns,calc}
\usepackage[round,authoryear]{natbib}
\usepackage[colorlinks=true,linkcolor=blue!60!black,citecolor=blue!60!black,urlcolor=blue!60!black]{hyperref}
\usepackage{caption}
\usepackage{microtype}
\hypersetup{
  pdftitle={What You Can't See Is What You Learn: Slot-Selective Evidence Masking Favors Compositional Generalization in Shared-Genome Language-Model Societies},
  pdfauthor={Narcis Marincat},
  pdfsubject={Slot-selective evidence masking and compositional generalization in shared-genome language-model societies}
}

\title{What You Can't See Is What You Learn:\\ Slot-Selective Evidence Masking Favors Compositional Generalization in Shared-Genome Language-Model Societies}
\author{Narcis Marincat\thanks{Independent researcher. Correspondence: \texttt{narcis.marincat.20@ucl.ac.uk}.}}
\date{August 2026}

\begin{document}
\maketitle

\begin{abstract}
Multi-module neural systems often expose every module to the full input. We test whether a slot-selective evidence-masking regime---restricting each module to its own evidence span---changes which solutions gradient-based training discovers. Four-cell societies share one frozen pretrained language model and one low-rank adapter, communicating only through two model-width continuous vectors in a fixed relay. On a prospectively sealed natural-language function-composition task, we train ten matched restricted/global pairs sharing initialization bytes, training order, token layout, positional geometry, trainable parameterization, and nominal computation budget; only the attention mask differs. Restricted-visibility societies outperform their globally visible twins by at least 20 percentage points at both depths in 9 of 10 pairs, with median paired advantages of 0.7648 and 0.6050. Cutting communication reduces every restricted society to chance, and in a post hoc collision-stratified analysis the depth-three advantage remains 0.558 on programs whose complete affine map never appeared in training. In six post hoc-selected restricted societies, packet interventions on correctly answered held-out episodes are consistent with approximately value-indexed relay states: same-value transplants preserve downstream behavior at 0.94--1.00 across all tested interfaces, destructive interventions collapse performance, and counterfactual packets redirect outputs toward the mathematically predicted answer. The sole high-performing global model also requires communication, but its same-value packets are not interchangeable across episodes. Thus restricted visibility is not necessary for composition. Under the tested seeds, streams, task world, and training budget, the masking regime strongly shifted which solutions training discovered: a post hoc mask crossover finds both arms mask-native, with no checkpoint retaining competence under the opposite evaluation mask, and the audited restricted interfaces were approximately value-indexed within their trained regime. Because the restricted mask both blocks foreign evidence and implicitly identifies each cell's assigned slot, attribution to evidence visibility alone awaits a role-marked control. The complete preregistered battery nevertheless formally fails because restricted-arm median depth-three accuracy is 0.6988, below the 0.70 floor. An earlier qualification cohort likewise yielded 0/10 complete passes: one model met every task-performance gate, but all ten failed ordinary-language preservation, confining the system to explicitly task-gated use.
\end{abstract}

\section{Introduction}
When a trainable module can inspect the complete program, whole-program lookup is an available solution. Restricting each module to one fragment removes that direct whole-program lookup route and may instead favor reusable local transformations connected by communication. This paper tests that hypothesis through a matched causal comparison of slot-selective evidence masking in a multi-module system built from a shared pretrained language model.

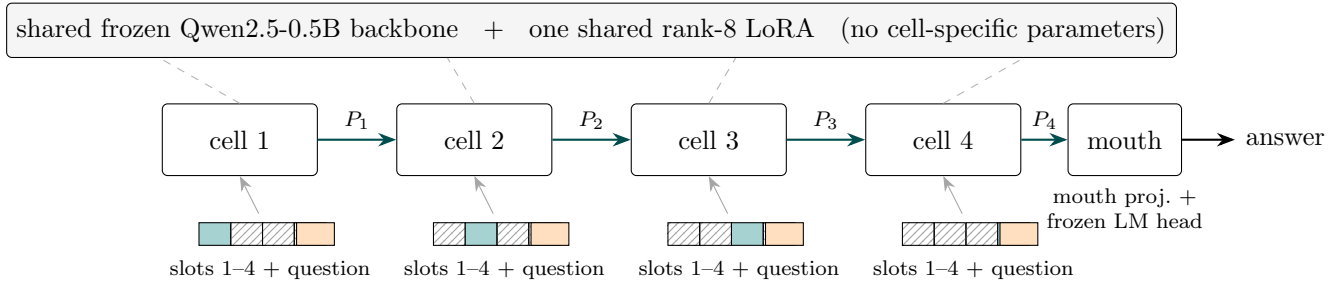
\begin{figure}[t]
\centering
\begin{tikzpicture}[
  cell/.style={draw, rounded corners=2pt, minimum width=2.05cm, minimum height=0.95cm, align=center, font=\small},
  slot/.style={draw, minimum width=0.42cm, minimum height=0.3cm, inner sep=0pt},
  pkt/.style={-{Stealth[length=2.4mm]}, thick, teal!60!black},
  lbl/.style={font=\scriptsize},
  x=1cm, y=1cm]
\node[draw, rounded corners=2pt, fill=gray!8, minimum width=12.6cm, minimum height=0.72cm, font=\small] (backbone) at (5.55,2.55)
  {shared frozen Qwen2.5-0.5B backbone \; + \; one shared rank-8 LoRA \; (no cell-specific parameters)};
\foreach \i/\x in {1/0.9, 2/4.0, 3/7.1, 4/10.2}{
  \node[cell] (c\i) at (\x,1.1) {cell \i};
  \draw[gray!60, dashed] (c\i.north) -- ($(backbone.south west)!{(\x+0.75)/12.6}!(backbone.south east)$);
}
\foreach \i/\x in {1/0.9, 2/4.0, 3/7.1, 4/10.2}{
  \foreach \j in {1,2,3,4}{
    \pgfmathsetmacro{\xs}{\x - 0.75 + 0.42*\j}
    \ifnum\i=\j
      \node[slot, fill=teal!35] (s\i\j) at (\xs, -0.15) {};
    \else
      \node[slot, pattern=north east lines, pattern color=gray!70] (s\i\j) at (\xs, -0.15) {};
    \fi
  }
  \node[slot, fill=orange!25, minimum width=0.5cm] (q\i) at (\x+1.0, -0.15) {};
  \node[lbl] at (\x+0.4,-0.62) {slots 1--4 + question};
  \draw[-{Stealth[length=2mm]}, gray!70] (\x+0.25,0.12) -- (c\i.south);
}
\node[cell, minimum width=1.5cm] (mouth) at (12.6,1.1) {mouth};
\node[lbl, below=0.02cm of mouth, align=center] {mouth proj.\ +\\ frozen LM head};
\draw[pkt] (c1.east) -- node[above, lbl, black] {$P_1$} (c2.west);
\draw[pkt] (c2.east) -- node[above, lbl, black] {$P_2$} (c3.west);
\draw[pkt] (c3.east) -- node[above, lbl, black] {$P_3$} (c4.west);
\draw[pkt] (c4.east) -- node[above, lbl, black] {$P_4$} (mouth.west);
\draw[-{Stealth[length=2.4mm]}, thick] (mouth.east) -- +(0.7,0) node[right, font=\small] {answer};
\end{tikzpicture}
\caption{\textbf{The society architecture and the treatment.} Four cells share one frozen backbone and one rank-8 LoRA. Cell 1 publishes $P_1$ from its assigned start-value span without incoming mail; cells 2--4 read $P_1$--$P_3$ and write $P_2$--$P_4$ (each packet: two 896-dimensional vectors, the only inter-cell pathway). Each cell receives an identical layout containing the shared question (orange) plus four evidence slots. In the restricted arm a cell's attention mask exposes only its own evidence slot (teal) and the question, leaving foreign slots (hatched) with exactly zero direct attention influence; in the global arm all four slots are readable. The attention mask is the sole implementation-level difference between twins; it jointly changes foreign-evidence access, role disambiguation, and active-context load (\S\ref{sec:limits}).}
\label{fig:arch}
\end{figure}

The experimental object is a \emph{society} of four cells sharing one frozen Qwen2.5-0.5B-Instruct genome \citep{qwen25} and one rank-8 adapter \citep{hu2022lora}. Cells communicate through learned continuous packets---two model-width vectors per hop, the only inter-cell information pathway---in a fixed relay ending in a frozen LM head plus a learned mouth projection. The task is ordered natural-language function composition over $\mathbb{Z}_{17}$; evaluation uses held-out ordered operator programs rendered in held-out phrasings. The treatment is a single attention-mask intervention. In the \emph{restricted} arm, every cell receives the same fixed four-slot input layout but may attend only to its assigned evidence span: foreign span tokens have exactly zero direct attention influence, and foreign information can reach that cell only through incoming packets. In the \emph{global} arm, all four spans are directly readable. Token layout, positional geometry, trainable parameterization, nominal token count, Transformer calls, training budget, initialization bytes, and per-pair training-example order are held fixed. Ten matched pairs---five initializations under two data orders---train for exactly 20{,}000 updates on a task instance generated and sealed before any training or outcome inspection, under preregistered gates and bit-exact machine verification (Figure~\ref{fig:arch}).

All ten globally visible twins fit the primitive-operation training set. On the held-out-phrasing single-operation diagnostic, accuracy ranges from 0.869 to 0.991. Nine of ten then display a memorization-without-generalization profile: all fit the depth-two training bank exactly (deterministic one-rendering-per-program/start-value evaluation of the full bank), while depth-three training-bank exact accuracy on a fixed 150-program subset ranges from 0.41 to 1.00 and held-out composition remains near chance at 0.05--0.11. All ten restricted-visibility twins achieve substantial held-out performance, ranging from 0.582 to 0.933 at depth three. One global twin (initialization 204, order 954) is the informative exception: it reaches 0.843 at depth three and its communication channel is causally necessary, but same-value packet transplants fail across episodes. By contrast, all six audited restricted-visibility societies learn approximately interchangeable carriers of the running intermediate value.

\paragraph{Contributions.}
\begin{enumerate}
\item \textbf{A matched causal comparison of slot-selective evidence masking.} We compare multi-module systems with identical architectures, inputs, positional geometry, trainable parameterization, nominal token count and Transformer calls, training budget, initialization bytes, and training-example streams, changing only whether each cell may attend to foreign evidence spans; in the restricted arm this masking also uniquely identifies each cell's assigned slot, so the treatment jointly manipulates evidence access and role disambiguation (\S\ref{sec:limits}). The comparison is prospectively sealed and evaluated with preregistered gates (\S\ref{sec:experiment}--\ref{sec:results}).
\item \textbf{A large effect that persists without composite-function collisions.} Median paired advantages are 0.7648 and 0.6050 at depths two and three. A post hoc collision-stratified analysis over nine complete pairs finds a median depth-three advantage of 0.558 on programs whose complete affine map never occurs in training, showing that exact lookup of a previously trained complete affine map does not explain the effect (\S\ref{sec:strat}).
\item \textbf{Mechanism tests using value-indexed packet transplants.} Beyond destructive channel interventions, we transplant natural packets between episodes according to the mathematical intermediate they represent. Across six audited restricted-visibility societies, same-value transplants preserve behavior at 0.94--1.00 across every tested interface, while counterfactual-value transplants steer outputs toward the mathematically predicted answer. The strongest model passes every absolute preregistered transplant gate. The same analysis separates the approximately interchangeable value code observed under the restricted masking regime from the non-interchangeable, episode-dependent code of the sole high-performing global model (\S\ref{sec:mechanism}).
\item \textbf{Complete outcome reporting and release of available artifacts.} We report two formal preregistered failures: an earlier qualification cohort in which 0/10 models passed the complete gate---although one passed every task-performance gate---and the present battery's 0.0012 miss on its absolute depth-three floor. We release all available final checkpoints, evaluations, audit scripts, preregistration materials, and incident records; one restricted checkpoint remains unavailable for collision-stratified analysis and is disclosed explicitly (\S\ref{sec:limits} and appendices).
\end{enumerate}

\section{Related Work}\label{sec:related}

\paragraph{Visibility controls in multi-agent systems.}
CoFlow \citep{zou2026coflow} supports full teammate visibility and an attention-masked agent-local mode within the same multi-agent architecture, providing a close visibility comparison in offline multi-agent reinforcement learning. \citet{bena2025} likewise compare shared and separate input pathways in controlled modular recurrent networks. These studies establish that module-level information access can be manipulated directly, but neither evaluates held-out program composition or causally audits a learned inter-module relay state. Our contribution is therefore not the first use of visibility masking in a multi-module system, but the conjunction of a matched slot-selective masking intervention, sealed compositional generalization, and value-indexed causal packet transplants.

\paragraph{Distributed observation and emergent codes.}
\citet{kaszynski2026} studies discrete compositional communication about latent physical properties inferred from frozen video features. In the current version, four-agent systems converge to near-perfect positional disentanglement in 80/80 seeds, while lower-agent-count and single-sender controls are markedly less reliable. The controls support an effect of distributed sender organization rather than total bandwidth or frame coverage. The comparison nevertheless changes agent count and message factorization; our experiment holds the four-cell architecture, packet bandwidth, initialization, nominal computation budget, and training stream fixed and changes only the attention mask at the implementation level; that mask jointly changes foreign-evidence access, role disambiguation, and active-context load. The wider emergent-communication literature \citep{lazaridou2017} established channel pressure as a shaper of codes: capacity and bandwidth constraints \citep{resnick2020}, the imperfect correlation of language compositionality with task generalization \citep{chaabouni2020}, co-adaptation control \citep{rita2022}, and partial observability shaping message content \citep{bosc2022}, typically with discrete channels and referential games rather than end-to-end latent relays on an LLM substrate.

\paragraph{Modularity, workspaces, and input separation.}
Recurrent Independent Mechanisms and Shared Global Workspace models study sparse, bandwidth-limited communication among neural modules \citep{goyal2021rims,goyal2022workspace}. \citet{bena2025} use controlled two-module recurrent networks to vary shared versus separate input pathways, sparse inter-module connectivity, resource constraints, and the timing and bandwidth of information flow. They show that structural modularity alone does not guarantee specialization and that specialization is favored by separable environmental features and resource constraints; they do not study systematic recomposition of held-out operator programs. The modular systematic-generalization literature \citep{lake2018,bahdanau2019,damario2021} shows that nominal modularity does not guarantee compositionality, and that scale can sometimes deliver it \citep{redhardt2025}---consistent with our framing of restriction as shifting basin probabilities rather than as a necessity claim.

\paragraph{Masking and context availability as inductive biases.}
AC-VLA \citep{peng2026acvla} combines instruction decomposition, dense subtask supervision, mixed training, and state-conditioned asymmetric masking of a shortcut visual input. Its ablations also isolate a substantial masking-only effect: adding wrist-view masking to raw-demonstration training improves spatial/goal OOD success from 35.5/46.6 to 47.3/67.0, while the full decomposition-plus-mask system reaches 64.2/73.3. This is a single-policy perceptual-shortcut intervention rather than a multi-module evidence-visibility or communication experiment. \citet{ma2026mask} show that task-structured additive attention-mask priors can persistently shift which extrapolating solution a Transformer learns on controlled Boolean and arithmetic tasks, although their masks are finite learnable biases rather than hard evidence-visibility boundaries. \citet{uzunoglu2026} show that abundant task-relevant train-time context can shift learning from parametric internalization toward contextual reliance, including through a redistribution of gradient pressure from feed-forward modules toward attention. \citet{ootani2026} compares information-matched symbolic, factored, shared, and entangled input routes in fully enumerable tiny Transformers, finding that input-pathway structure affects few-shot binding---a controlled single-model precedent for input routing shaping the learned solution. Together these works support the broader hypothesis that information availability can redirect optimization, but none studies a matched multi-cell latent relay in which a slot-selective attention mask is the only implementation-level difference and the communicated state is causally audited.

\paragraph{Multi-agent LLM systems and latent channels.}
Inference-time LLM societies typically exchange natural-language messages, often with substantial context overlap \citep{li2023camel,wu2023autogen,du2024debate}, while recent systems communicate through embeddings, hidden states, or KV caches. Aggregate gains and destructive message ablations do not by themselves identify whether a receiver uses example-specific sender information. \citet{zhang2026latent} intervene at the latent-message boundary using no-message, other-example, self-generated, and current-example messages; \citet{cheng2026kv} similarly compare native, deranged, zeroed, and moment-matched KV relays. Their findings show that a large channel effect can survive replacement by an answer-irrelevant message. StateBridge \citep{peng2026statebridge} aligns sender hidden states to a receiver's input space using a training-free closed-form orthogonal transformation with norm calibration and vocabulary anchoring; it addresses cross-agent latent-space compatibility and portability rather than learned protocol formation, evidence visibility, or causal semantic interchange. We therefore treat destructive interventions as necessary controls and ground the mechanism claim in natural same-value and counterfactual-value packet transplants.

\paragraph{Causal abstraction and interchange interventions.}
Causal abstraction treats a mechanistic hypothesis as an alignment between high-level causal variables and low-level neural states, tested by asking whether interchange interventions produce the corresponding counterfactual effects \citep{geiger2021causal,geiger2025jmlr}. Distributed Alignment Search extends this framework to variables represented in non-standard distributed subspaces \citep{geiger2024das}. Our packet transplants apply the same criterion at an explicit communication boundary, where the proposed high-level variable---the running $\mathbb{Z}_{17}$ value---is known exactly.

\paragraph{Reusable causal interfaces.}
Hidden APIs in Language Models \citep{ma2026hidden} uses forked future operations, architecture competition, role-aligned transplantation, locality controls, and causal mediation to identify reusable interfaces within monolithic language models. It is the closest precedent for our claim that causally equivalent internal states should remain interchangeable under different downstream consumers. We instead study a state learned expressly for inter-cell communication and ask how evidence visibility changes the probability and form of that protocol.

\paragraph{Regularization by information removal.}
Dropout and modality dropout remove information stochastically \citep{hinton2012,srivastava2014,neverova2016}, while information-bottleneck methods constrain representational capacity \citep{tishby2000}. Our treatment instead imposes a stable semantic ownership boundary: each cell may access one assigned evidence span, and task-essential foreign evidence can influence it only through the learned packet channel. Because \emph{privacy} in federated learning \citep{mcmahan2017} usually denotes formal or operational data-protection properties that we do not study, we use \emph{restricted evidence visibility} or \emph{evidence partitioning} throughout.

\paragraph{Novelty, stated exactly.}
Prior work separately studies restricted information access, compositional emergent communication, latent inter-agent messages, and causal interchange interventions. To our knowledge, this is the first study to combine a matched intervention on a slot-selective attention mask in a learned inter-cell latent relay, prospectively sealed held-out compositional programs, and causal interchange tests establishing the communicated intermediate variable.

\section{Societies of LLM cells}\label{sec:arch}

\paragraph{Architecture.}
A society comprises four cells and one readout. All cells share a frozen Qwen2.5-0.5B-Instruct backbone and one rank-8 LoRA applied to its attention projections; there are no cell-specific parameters or learned cell identities. Execution is staged: cell 1 publishes from the start-value span without incoming mail, and cells 2--4 each consume the predecessor packet once while applying their assigned operation or structural forwarder. Two shared bias-free projections implement packet reading and writing, and a third bias-free projection implements mouth injection. The reader maps an incoming normalized $2\times896$ packet into two pseudo-token embeddings appended to the cell input; the writer maps the corresponding final hidden states into a packet residual. Packets update as
\[
P_{\mathrm{out}}=\operatorname{RMSNorm}\left(P_{\mathrm{in}}+0.05\,\Delta P\right).
\]
The mouth forms
\[
h_{\mathrm{final}} = h_{\mathrm{base}} + \beta\, W_{\mathrm{mouth}}(P_4), \qquad \beta = 0.003,
\]
at the answer position, where $h_{\mathrm{base}}$ is the adapter-disabled frozen-base representation and $\beta$ is calibrated and frozen before training; the frozen LM head then produces the output. For the 17 task labels, training and task evaluation use contextually residualized logits,
\[
\tilde z_j = z_j(h_{\mathrm{final}}) - z_j(h_{\mathrm{base}}),
\]
with the adapter disabled on the base path; ordinary-language preservation is evaluated separately on the uncorrected full-vocabulary logits. The system has approximately 4.32M trainable parameters. Training uses answer-token cross-entropy together with a packet-level auxiliary classifier whose weight is annealed to zero by update 10{,}000; all reported evaluations disable the auxiliary head and use only the frozen LM head plus the learned mouth projection.

\paragraph{Task and curriculum.}
An episode contains an initial value in $\mathbb{Z}_{17}$ and zero to three natural-language operation spans drawn from 12 affine bijections. The answer is encoded by one of 17 tokenizer-verified single-token labels. Training samples identity episodes with probability 0.10, single-operation episodes with probability 0.25, and two- and three-operation episodes with probability 0.325 each. Identity and single-operation episodes were selected as essential curriculum atoms in a pre-battery development ladder rather than introduced after the paired results were observed. Evaluation uses held-out ordered programs, held-out phrasings, and separately reported composite-function strata. Operator tables, program splits, phrasing grammars, and answer-token mappings are generated and hashed before training.

\paragraph{Treatment: restricted versus global evidence visibility.}
Both arms receive the same fixed four-slot token sequence with identical padding, positions, packet slots, and Transformer calls. In the restricted arm, a cell's attention mask exposes only its assigned evidence slot and the shared question; in the global arm, the same cell may attend to all four evidence slots. The attention mask is the sole implementation-level treatment; in the restricted arm it both blocks foreign evidence and uniquely identifies the cell's assigned slot (\S\ref{sec:limits}).

\section{The paired-visibility experiment}\label{sec:experiment}

Ten twin pairs---five initializations under two data-order streams---train for exactly 20{,}000 updates on the sealed instance; each pair shares initialization bytes and its ordered example stream, verified through running stream hashes. Machines pass two bit-exact reproduction gates before contributing. The paired-effect, absolute-performance, communication-necessity, and packet-intervention thresholds were fixed before training; collision-stratified rescoring and inferential tests were added after outcome inspection and are labeled post hoc. Evaluation uses final checkpoints only. We additionally train three staged centralized-scan models with cumulative evidence visibility as secondary comparators. An anonymized preregistration, protocol history, and incident ledger are included in the supplementary material.

\paragraph{Statistical unit and analysis.}
The preregistered gates count ten complete paired training trajectories. Because each model initialization is reused under two data orders, inferential uncertainty is clustered by initialization, yielding five initialization strata. We report all ten paired differences, the five initialization-level mean differences, cluster-bootstrap intervals, and an exact sign-flip analysis over the five initialization-level means as post hoc descriptive inference. Evaluation examples are used only to estimate within-checkpoint accuracy and are not treated as independent training replicates.

\section{Results}\label{sec:results}

\subsection{The paired effect}\label{sec:paired}
\begin{figure}[t]
\centering
\includegraphics[width=\textwidth]{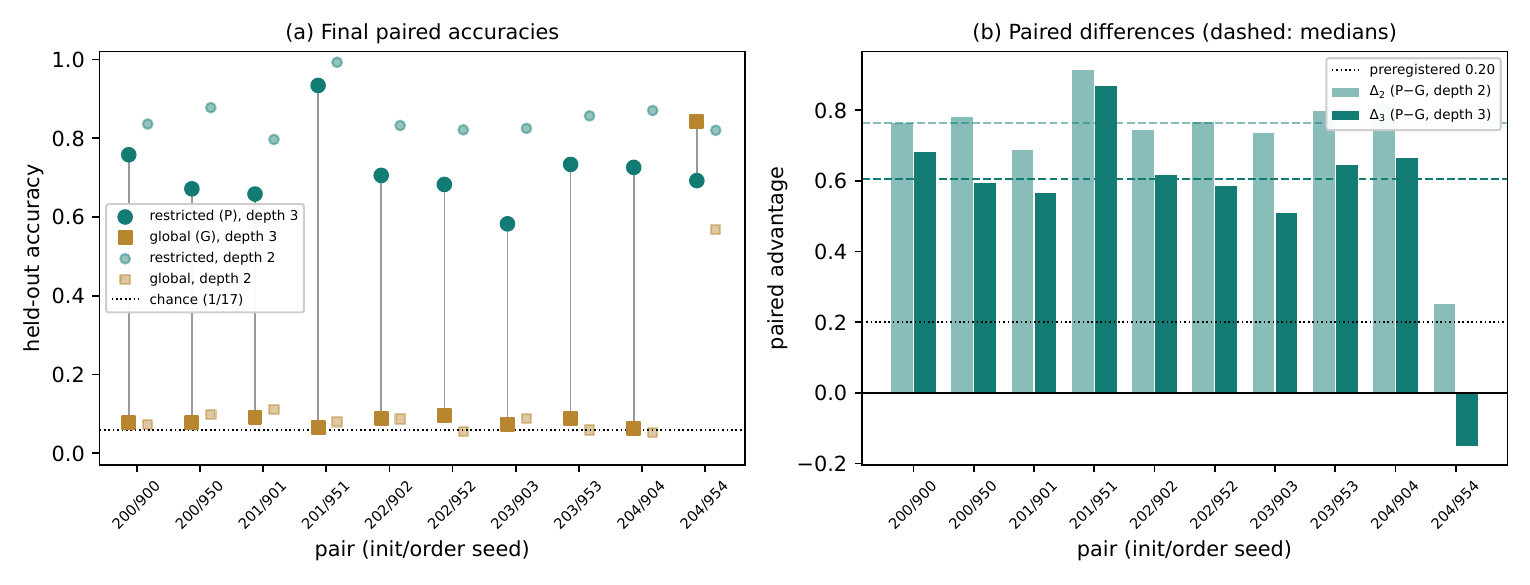}
\caption{\textbf{The paired masking-regime effect.} (a) Final held-out accuracies for all ten twin pairs; vertical grey segments connect each pair at depth three. (b) Paired advantages $\Delta_2$ and $\Delta_3$ per pair with medians (dashed) and the preregistered 0.20 per-pair threshold (dotted). Pair 204/954 is the globally visible exception discussed in \S\ref{sec:exception}.}
\label{fig:pairs}
\end{figure}

(Figure~\ref{fig:pairs}: final paired accuracies and paired differences; Table~\ref{tab:pairs}: all ten pairwise results.) Restricted-visibility societies exceed their global twins by at least 0.20 at both depths in 9 of 10 pairs; median paired advantages are 0.7648 at depth two and 0.6050 at depth three. Every restricted society is communication-dependent: severing all packets yields exactly chance performance, $1/17$, in all ten. Nine global twins display the memorization-without-generalization profile: all fit the primitive-operation training set and the depth-two training bank exactly, depth-three training-bank exact accuracy on the fixed first-150-program subset ranges from 0.41 to 1.00, and held-out accuracy remains at 0.05--0.11. The preregistered gate decomposition is shown in Table~\ref{tab:gates}. Every component passes except the absolute restricted-arm depth-three floor, whose observed median is 0.6988 rather than 0.70; the complete conjunction therefore formally fails.

As post hoc descriptive inference clustered by initialization (\S\ref{sec:experiment}), all five initialization-level mean differences are positive at both depths (depth two: 0.7715, 0.7996, 0.7555, 0.7672, 0.5350; depth three: 0.6367, 0.7176, 0.6020, 0.5774, 0.2563); post hoc percentile cluster-bootstrap intervals over the five initialization strata (nominally 95\%, reported descriptively because interval coverage with five clusters is uncertain) for the mean paired advantage are $[0.628, 0.784]$ at depth two and $[0.397, 0.669]$ at depth three, and the exact two-sided five-stratum sign-flip $p$-value is 0.0625 at each depth---the smallest value attainable with five strata.

\begin{table}[t]
\centering\small
\caption{\textbf{Preregistered gate decomposition.} Thresholds were frozen before training; the complete conjunction determines the preregistered verdict.}
\label{tab:gates}
\begin{tabular}{lr}
\toprule
Preregistered component & Result \\
\midrule
$P-G\ge 0.20$ at both depths & 9/10 pairs --- pass \\
Median $\Delta_2\ge 0.25$ & 0.7648 --- pass \\
Median $\Delta_3\ge 0.25$ & 0.6050 --- pass \\
Restricted-arm all-cut communication & 10/10 at $1/17$ --- pass \\
No restricted run below 0.40 & pass \\
Restricted median depth two $\ge 0.70$ & 0.8339 --- pass \\
Restricted median depth three $\ge 0.70$ & 0.6988 --- \textbf{fail} \\
\midrule
Complete preregistered conjunction & \textbf{fail} \\
\bottomrule
\end{tabular}
\end{table}

\begin{table}[t]
\centering\small
\caption{\textbf{All ten pairwise results.} Held-out accuracy at depths two and three for the restricted (P) and global (G) twin of each pair, paired differences, and all-packets-cut accuracy at depth three (chance $=1/17\approx0.0588$).}
\label{tab:pairs}
\begin{tabular}{cccccccccc}
\toprule
init & order & P$_2$ & P$_3$ & G$_2$ & G$_3$ & $\Delta_2$ & $\Delta_3$ & P all-cut & G all-cut \\
\midrule
200 & 900 & 0.8358 & 0.7578 & 0.0723 & 0.0770 & +0.7635 & +0.6808 & 0.0588 & 0.0600 \\
200 & 950 & 0.8775 & 0.6713 & 0.0980 & 0.0787 & +0.7795 & +0.5926 & 0.0588 & 0.0701 \\
201 & 901 & 0.7966 & 0.6581 & 0.1103 & 0.0917 & +0.6863 & +0.5664 & 0.0588 & 0.0679 \\
201 & 951 & 0.9926 & 0.9338 & 0.0797 & 0.0650 & +0.9129 & +0.8688 & 0.0588 & 0.0674 \\
202 & 902 & 0.8321 & 0.7054 & 0.0870 & 0.0880 & +0.7451 & +0.6174 & 0.0588 & 0.0588 \\
202 & 952 & 0.8211 & 0.6824 & 0.0551 & 0.0958 & +0.7660 & +0.5866 & 0.0588 & 0.0512 \\
203 & 903 & 0.8248 & 0.5824 & 0.0882 & 0.0725 & +0.7366 & +0.5099 & 0.0588 & 0.0684 \\
203 & 953 & 0.8566 & 0.7333 & 0.0588 & 0.0885 & +0.7978 & +0.6448 & 0.0588 & 0.0657 \\
204 & 904 & 0.8701 & 0.7257 & 0.0527 & 0.0623 & +0.8174 & +0.6634 & 0.0588 & 0.0650 \\
204 & 954 & 0.8199 & 0.6922 & 0.5674 & 0.8431 & +0.2525 & -0.1509 & 0.0588 & 0.0569 \\
\bottomrule
\end{tabular}
\end{table}

The three fresh-world staged centralized models score 1.000, 0.993, and 0.098 at depth three, indicating high between-seed variability rather than uniform incapacity. These models were not paired to the masking cohort, were not compute-matched to the four-cell society, and were trained on a different GPU class. On the earlier qualification-world split, three atom-trained flat one-call comparators scored 0.117--0.200 at depth two and 0.100--0.267 at depth three, while all three staged controls failed; a single staged development run on another split succeeded. We therefore treat all centralized results as contextual baselines rather than as the causal identification of the masking-regime effect. Across the earlier restricted-only world and the paired world, all twenty restricted-visibility societies exceeded 0.58 at depth three.

\subsection{The globally visible exception}\label{sec:exception}
The global twin of initialization 204 under order 954 reaches 0.843 at depth three, exceeding its restricted sibling, and also collapses under packet deletion. Global visibility is therefore compatible with a high-performing, communication-dependent strategy in this architecture. In the present ten-pair cohort, however, only one global trajectory reaches that regime. Section~\ref{sec:mechanism} shows that its learned packet interface differs from the approximately value-indexed interface observed in the audited restricted models.

\subsection{Post hoc collision-stratified scoring}\label{sec:strat}
Many held-out programs share their complete affine map with a training program, so a function-level lookup strategy could in principle score well without executing the unseen operator sequence. We therefore stratify the available final checkpoints by whether the composite function itself appeared in training. Collision-stratified results are available for 19 of 20 checkpoints, yielding nine complete restricted/global pairs; one restricted checkpoint remains archive-pending. On map-novel depth-three programs, the median paired advantage remains $+0.558$, compared with $+0.634$ on the map-redundant stratum. Restricted models score 0.553--0.913 on map-novel programs, ordinary global models remain near chance at 0.049--0.094, and the 204/954 global model reaches 0.828. Composite-map collisions therefore reduce the benchmark's discriminative purity in principle but do not explain either the observed paired masking-regime effect or the existence of compositional computation. We report the small residual redundant-versus-novel advantage separately and recommend collision-free future task worlds.

\section{Mechanism: value-indexed packet transplants}\label{sec:mechanism}

Standard destructive audits (edge deletion, cross-example shuffle, approximately norm-matched noise) establish channel necessity but not content. The audit's cross-example shuffle is interface-specific and draws natural donor packets within the audited episode pool; it is distinct from the all-interface batch-shuffle diagnostic reported in Table~\ref{tab:eval}, which rolls committed packets across the evaluation batch at all three interfaces simultaneously. Our transplants are indexed by the semantic intermediate value: a donor packet from a different episode replaces the recipient's packet at an interface, with the donor chosen so the running value either matches (same-value) or differs with a known target (counterfactual).

After behavioral evaluation, we selected six restricted-visibility societies spanning the observed performance range for packet auditing. Across these six societies, among held-out episodes that each model answered correctly, natural packets from different episodes are nearly interchangeable when they denote the same mathematical intermediate: same-value transplant accuracy is 0.94--1.00 across all tested interfaces. Deleting packets or replacing them with approximately norm-matched noise reduces performance to at most 0.11, and cross-example shuffling scores 0.07--0.11. Counterfactual transplants redirect outputs toward the mathematically predicted answer, with fidelity ranging from 0.74 to 1.00 and tracking each model's intact task competence; the strongest model passes every absolute preregistered transplant gate. These results provide evidence that, on correctly answered held-out episodes, the audited restricted societies repeatedly implement an approximately value-indexed causal relay, although only the strongest model supports the phrase \emph{exact causal abstraction} under all absolute thresholds.

The 204/954 global twin dissociates sharply. Packet deletion collapses its performance, but same-value transplants succeed on only 0.12--0.25 of cases; its communication state is therefore strongly episode- or context-dependent rather than reducible to the intermediate value alone. Among the audited models, these interventions provide evidence that the complete masking regime affected not only how often a successful strategy was learned but also the form of the learned interface: all six audited restricted societies exhibit an approximately interchangeable value-indexed code, whereas the sole high-performing global model uses a non-interchangeable, episode-entangled code, operationally defined by the failure of same-value packets to remain causally interchangeable across episodes. Because the global arm produced only one high-performing model, this is a mechanistic case study rather than an estimate of how frequently global visibility produces each code class. We hypothesize that blocking foreign spans pressures the receiver toward a more self-sufficient packet representation. However, the same mask also supplies role disambiguation and reduces active-context load, so the causal ingredient behind the code-form difference remains unresolved pending the role-marked control.

\paragraph{Post hoc train-mask $\times$ evaluation-mask crossover (added in v3).} We reevaluated every available final checkpoint under the opposite arm's attention mask using a single evaluator override; all native-mask cells reproduced the original six evaluation metrics exactly. Under the restricted evaluation mask, all ten globally trained models remained at chance at depth three (median 0.0583, range 0.0544--0.0669), including the sole high-performing global model, which fell from 0.8431 under its native mask to 0.0581. Conversely, all nine recoverable restricted-trained checkpoints fell to chance under the global evaluation mask (median 0.0615, range 0.0539--0.0694). Depth-two results showed the same pattern. These crossed evaluations are distribution shifts and therefore do not separately identify the effects of role cues, foreign semantic evidence, or active-context load. They provide no support for the simple rescue hypothesis that the global-trained checkpoints already contained a restricted-mask-compatible procedure whose native errors arose only from acute exposure to additional spans. Instead, both arms learned mask-native procedures: restricted-trained solutions are not robust to unmasking, while the successful global solution depends on the full-context regime in which it was trained. The sole globally visible success is fully native-context dependent, consistent with the packet-transplant evidence that its computation is context-dependent, although the crossover perturbs every cell's input and cannot attribute the collapse specifically to the packet code. For the restricted models, the collapse under unmasking does not negate their generalization to unseen programs under the architecture in which they were trained, but it rules out mask-independent algorithmicity: any phrase such as ``reusable relay'' means reusable across held-out programs and the tested near-transfer family within the restricted attention regime. Crossover artifacts are released alongside the original evaluations.

\section{Limitations and scope}\label{sec:limits}

The causal masking-regime comparison concerns one architecture family, one templated synthetic task family, one prospectively sealed paired task world, and a fixed 20{,}000-update budget. The ten paired trajectories arise from five model initializations evaluated under two data orders, so they are not ten fully independent initialization draws; uncertainty analyses are therefore clustered by initialization. A prior restricted-only world provides behavioral context but is not an independent paired replication of the visibility effect. Restricted visibility is not necessary for composition: one global model is a constructive counterexample. The supported claim is instead that, under the tested seeds, streams, task world, and budget, the masking regime strongly shifted the observed frequency and form of the learned solution. Nor is ours the first multi-agent local/global visibility comparison \citep[e.g.,][]{zou2026coflow}; the contribution is the tightly paired, mechanism-audited form of the comparison.

\paragraph{Role-cue limitation.} Although the attention mask is the only implementation-level difference between arms, it changes three coupled aspects of the cell's information state: foreign-evidence access, role or ownership information, and active-context load. In the restricted arm, making only one evidence slot readable also uniquely identifies the cell's assigned slot; in the global arm, all slots are readable and there is no explicit current-cell or own-slot marker. The present comparison therefore identifies the effect of the complete masking regime, jointly including foreign-evidence access, role disambiguation, and active-context load. The post hoc crossover above rejects a simple test-time rescue but, being a distribution shift, does not resolve this confound, because the mask is part of the training regime. A fresh comparison in which both arms receive an explicit role marker is required to isolate foreign-evidence access from role disambiguation.

\paragraph{Held-out monitoring disclosure.} A fixed subset of the held-out diagnostic banks was evaluated every 500 updates for monitoring. These probes did not enter the loss, stopping rule, checkpoint selection, run selection, or formal verdict, all of which were fixed in advance. The diagnostic family was therefore held out from gradient training but was not blind to the experimenters.

The complete preregistered battery formally fails because restricted-arm median depth-three accuracy is 0.6988 rather than the required 0.70. An earlier qualification cohort also passes 0/10 complete gates; one model meets all task-performance criteria, but all ten fail the ordinary-language preservation gate, with approximately 50--61 percentage points of top-1 regression. The present system is therefore an explicitly task-gated specialist, not an always-on zero-regression language-model retrofit.

The modular arithmetic primitives are not native capabilities of the frozen base model. Training explicitly includes identity and single-operation curriculum atoms in addition to multi-operation programs. Thus, pretraining supplies a shared linguistic representation space, while the arithmetic primitives, packet protocol, and composition procedure are learned from the synthetic curriculum.

The inter-cell channel is bounded but not minimal: each message contains two 896-dimensional continuous vectors. Six of ten restricted models were selected for packet audits to span the performance range, and only one high-performing global model was available for a code-type comparison. The claim that the restricted masking regime favors a value-indexed interface is therefore supported in the audited sample---on correctly answered episodes---but not yet quantified as a population frequency. Four restricted checkpoints remain unaudited.

Restricted and global twins are matched in token layout, parameters, Transformer calls, and training streams, but the secondary centralized controls are neither pair-matched nor compute-matched to the four-cell society. The experiment establishes a masking-regime effect within the society architecture; it does not establish computational efficiency or superiority to every centralized architecture.

Collision-stratified analysis is available for 19 of 20 final checkpoints. Composite-function collisions do not explain the observed effect, but they make the aggregate benchmark less discriminative than a collision-free construction. One restricted checkpoint remains unavailable for that analysis and is disclosed. Finally, the study does not establish cross-family packet universality, free-form language reasoning, dynamic routing, event-driven quiescence, long-context utility, or any claim about consciousness. Evidence ownership, relay topology, task engagement, and the number of communication stages are externally specified rather than learned. Collision-stratified rescoring and the choice of models for packet auditing are post hoc descriptive analyses, although their intervention criteria were fixed independently of the resulting scores.

\bibliographystyle{plainnat}
\bibliography{references}

\appendix

\section{Protocol details}\label{app:protocol}
\paragraph{Task instance and sealing.} The paired task world is generated from three seeds fixed and hashed before any training: operator seed 6011, split seed 2203 (split hash \texttt{94e506d408b6def1}), and phrasing-grammar seed 7717 (grammar hash \texttt{def14eb4182e2949}). Twelve affine bijections over $\mathbb{Z}_{17}$ form the operator set; programs are ordered operator sequences of length 0--3. Program banks are split so that held-out depth-two and depth-three programs never occur as ordered sequences in training; a bank-intersection audit verifies the separation. Held-out programs are rendered in held-out phrasings drawn from a templated grammar with four surface rotations per operator.

\paragraph{Evaluation sets.} The depth-two held-out set contains 12 programs and the depth-three set 60 programs; each program is evaluated over all 17 start values and 4 phrasing rotations, giving 816 and 4{,}080 scored episodes per checkpoint respectively. Task training and exact-match evaluation use the 17 residualized label logits defined in \S\ref{sec:arch}; ordinary-language preservation uses uncorrected full-vocabulary logits. Chance task accuracy is $1/17\approx0.0588$.

\paragraph{Layout constants.} Question length $T_q=54$ tokens, span length $T_s=33$ tokens, four evidence slots per episode, and $M_P=2$ packet pseudo-token slots yield 132 evidence-slot tokens and 188 total tokens per adapter-active cell call. Each episode uses four adapter-active cell calls plus one adapter-disabled, question-only mouth-base call. Both arms use a fixed $2\times16$ gradient-accumulation schedule; batch composition and ordering are identical within each pair (verified by running stream hashes).

\paragraph{Preregistered gates.} (i) $\Delta\ge0.20$ at both depths in at least 8 of 10 pairs; (ii) median $\Delta_2\ge0.25$; (iii) median $\Delta_3\ge0.25$; (iv) all-cut communication at $\le0.11$ in at least 8 restricted models (observed: exactly chance in 10/10); (v) no restricted run below 0.40; (vi) restricted median $\ge0.70$ at both depths. Machine admission requires bit-exact reproduction of two golden fingerprints (restricted \texttt{c7fcb21aad26951b}, global \texttt{b7c6157b1b3b79af}) computed over internal state after a fixed training prefix.

\section{Full result tables}\label{app:tables}

\begin{table}[h]
\centering\scriptsize
\caption{\textbf{Per-model evaluation diagnostics, all twenty finals.} $\ell_0$: identity accuracy; $\ell_1$: single-operation (held-out phrasing) accuracy; all-cut: all packets severed (depth three); best cell: best single-cell-only accuracy; batch shuffle: all-interface batch packet shuffle (committed packets rolled across the evaluation batch at all three interfaces simultaneously --- a broad diagnostic distinct from the audit's interface-specific cross-example shuffle, Table~\ref{tab:audit}); train3: depth-three training-bank exact accuracy on the first 150 training programs. Depth-two training-bank exact accuracy is 1.000 for every model and is omitted.}
\label{tab:eval}
\begin{tabular}{ccccccccccc}
\toprule
arm & init & order & $\ell_0$ & $\ell_1$ & depth 2 & depth 3 & all-cut & best cell & batch shuffle & train3 \\
\midrule
G & 200 & 900 & 1.000 & 0.9914 & 0.0723 & 0.0770 & 0.0600 & 0.0600 & 0.0691 & 1.00 \\
G & 200 & 950 & 1.000 & 0.9816 & 0.0980 & 0.0787 & 0.0701 & 0.0701 & 0.0735 & 0.78 \\
G & 201 & 901 & 1.000 & 0.9755 & 0.1103 & 0.0917 & 0.0679 & 0.0679 & 0.0809 & 0.42 \\
G & 201 & 951 & 1.000 & 0.9743 & 0.0797 & 0.0650 & 0.0674 & 0.0674 & 0.0681 & 0.41 \\
G & 202 & 902 & 1.000 & 0.9265 & 0.0870 & 0.0880 & 0.0588 & 0.0588 & 0.0814 & 0.43 \\
G & 202 & 952 & 1.000 & 0.9534 & 0.0551 & 0.0958 & 0.0512 & 0.0512 & 0.0880 & 0.99 \\
G & 203 & 903 & 1.000 & 0.8689 & 0.0882 & 0.0725 & 0.0684 & 0.0684 & 0.0686 & 0.86 \\
G & 203 & 953 & 1.000 & 0.9632 & 0.0588 & 0.0885 & 0.0657 & 0.0657 & 0.0782 & 0.48 \\
G & 204 & 904 & 1.000 & 0.9620 & 0.0527 & 0.0623 & 0.0650 & 0.0650 & 0.0549 & 0.51 \\
G & 204 & 954 & 1.000 & 0.8995 & 0.5674 & 0.8431 & 0.0569 & 0.0569 & 0.6627 & 1.00 \\
P & 200 & 900 & 1.000 & 0.9277 & 0.8358 & 0.7578 & 0.0588 & 0.0686 & 0.5784 & 1.00 \\
P & 200 & 950 & 1.000 & 0.8909 & 0.8775 & 0.6713 & 0.0588 & 0.0686 & 0.5186 & 1.00 \\
P & 201 & 901 & 1.000 & 0.8860 & 0.7966 & 0.6581 & 0.0588 & 0.0686 & 0.5123 & 1.00 \\
P & 201 & 951 & 1.000 & 0.9926 & 0.9926 & 0.9338 & 0.0588 & 0.0686 & 0.6900 & 1.00 \\
P & 202 & 902 & 1.000 & 0.9020 & 0.8321 & 0.7054 & 0.0588 & 0.0686 & 0.5353 & 1.00 \\
P & 202 & 952 & 1.000 & 0.8897 & 0.8211 & 0.6824 & 0.0588 & 0.0686 & 0.5201 & 1.00 \\
P & 203 & 903 & 1.000 & 0.8591 & 0.8248 & 0.5824 & 0.0588 & 0.0686 & 0.4532 & 1.00 \\
P & 203 & 953 & 1.000 & 0.9179 & 0.8566 & 0.7333 & 0.0588 & 0.0686 & 0.5586 & 1.00 \\
P & 204 & 904 & 1.000 & 0.9142 & 0.8701 & 0.7257 & 0.0588 & 0.0686 & 0.5554 & 1.00 \\
P & 204 & 954 & 1.000 & 0.8946 & 0.8199 & 0.6922 & 0.0588 & 0.0686 & 0.5284 & 1.00 \\
\bottomrule
\end{tabular}
\end{table}

\begin{table}[h]
\centering\small
\caption{\textbf{Post hoc collision-stratified depth-three scores per pair.} Map-novel: composite affine map absent from training (13 of 60 programs; 884 scored episodes per checkpoint); map-redundant: map present (47 programs; 3{,}196 episodes). At depth two, 3 of 12 programs are map-novel. One restricted checkpoint (init 202, order 952) is archive-pending and disclosed.}
\label{tab:strat}
\begin{tabular}{ccccccc}
\toprule
init & order & P novel & G novel & $\Delta$ novel & P redundant & G redundant \\
\midrule
200 & 900 & 0.7104 & 0.0667 & +0.6437 & 0.7710 & 0.0798 \\
200 & 950 & 0.5905 & 0.0532 & +0.5373 & 0.6937 & 0.0857 \\
201 & 901 & 0.6131 & 0.0939 & +0.5192 & 0.6705 & 0.0911 \\
201 & 951 & 0.9129 & 0.0690 & +0.8439 & 0.9396 & 0.0638 \\
202 & 902 & 0.6369 & 0.0792 & +0.5577 & 0.7243 & 0.0904 \\
202 & 952 & \multicolumn{5}{c}{restricted checkpoint archive-pending (disclosed)} \\
203 & 903 & 0.5532 & 0.0792 & +0.4740 & 0.5904 & 0.0707 \\
203 & 953 & 0.6776 & 0.0814 & +0.5962 & 0.7487 & 0.0904 \\
204 & 904 & 0.6697 & 0.0486 & +0.6211 & 0.7412 & 0.0660 \\
204 & 954 & 0.6324 & 0.8281 & -0.1957 & 0.7087 & 0.8473 \\
\bottomrule
\end{tabular}
\end{table}

\begin{table}[h]
\centering\scriptsize
\caption{\textbf{Packet-audit summary, eight audited models.} Ranges span the three relay interfaces (cell1$\to$2, 2$\to$3, 3$\to$4). Each audit begins with 180 held-out episodes and conditions intervention analyses on the subset the model answers correctly; the resulting per-model denominators $n$ are shown in the table (the $n=12$ global control's estimates are correspondingly imprecise), and exact values are recorded in the released audit JSONs. Base: intact accuracy on the audited episode pool. Del/noise: worst (highest) of packet deletion and approximately norm-matched noise. Shuffle: cross-example packet shuffle (its preregistered gate accepts either $\le0.11$ or $\ge$30 points below intact accuracy). Span-cf: counterfactual evidence-span rewrite control. Gate marks are the preregistered absolute audit gates.}
\label{tab:audit}
\begin{tabular}{lcccccccc}
\toprule
model & $n$ & base & same-value & counterfactual & span-cf & del/noise & shuffle & all gates \\
\midrule
G:m200:o950 & 12 & 0.067 & 0.00--0.50 & 0.00--0.33 & 0.25--0.50 & 0.250 & 0.25--0.33 & no \\
G:m204:o954 & 155 & 0.861 & 0.12--0.25 & 0.15--0.26 & 0.70--0.91 & 0.071 & 0.06--0.11 & no \\
P:m200:o950 & 114 & 0.633 & 0.96--1.00 & 0.91--0.98 & 0.86--0.88 & 0.105 & 0.04--0.08 & no \\
P:m201:o951 & 160 & 0.889 & 0.99--1.00 & 0.98--1.00 & 0.97--0.98 & 0.062 & 0.05--0.07 & yes \\
P:m203:o903 & 96 & 0.533 & 0.96--1.00 & 0.74--0.90 & 0.73--0.82 & 0.073 & 0.06--0.07 & no \\
P:m203:o953 & 125 & 0.694 & 0.94--1.00 & 0.86--0.92 & 0.81--0.86 & 0.096 & 0.08--0.11 & no \\
P:m204:o904 & 115 & 0.639 & 0.97--1.00 & 0.89--0.95 & 0.83--0.89 & 0.078 & 0.07--0.09 & no \\
P:m204:o954 & 111 & 0.617 & 0.98--1.00 & 0.90--0.96 & 0.77--0.95 & 0.108 & 0.06--0.07 & no \\
\bottomrule
\end{tabular}
\end{table}

\paragraph{Donor selection rules.} For each audited interface, donor packets are harvested from natural runs of \emph{different} episodes at the same interface position. Same-value donors carry an identical running intermediate; counterfactual donors carry a different, known value whose mathematically implied final answer defines the steering target. Donors are matched only by value, never by episode, phrasing, or start value; no packet is synthesized or optimized. The span-counterfactual control rewrites the evidence text instead, verifying that packet steering matches text-level causal ground truth.

\paragraph{Cluster-bootstrap intervals.} As a post hoc descriptive analysis, we resample the five initialization strata with replacement ($10^5$ draws, seed 20260820, percentile method). The nominal 95\% percentile intervals for the mean paired advantage are $[0.628, 0.784]$ at depth two and $[0.397, 0.669]$ at depth three; because there are only five clusters, their frequentist coverage is uncertain and they should be read descriptively. The exact two-sided sign-flip $p$-value over the five stratum means is $0.0625$ at each depth, the smallest attainable value with five strata.

\section{Development history}\label{app:history}
The program preceding the paired battery contributed three results that shaped the present design. First, an earlier recurrent-genome program established the packet-transplant methodology and reported its negative results: under a competitive gating mechanism, learning signal attenuated with relay depth, and the preregistered battery failed; replacing the gate restored deep relaying (fidelity by depth approximately 1.00/1.00/0.86/0.75). Second, a solvability ladder showed that, under the tested protocol, multi-operation-only training fit training programs but failed standalone primitive execution and held-out composition; adding identity and single-operation curriculum atoms yielded systematic generalization in the staged development control, motivating their inclusion in the frozen paired protocol. Third, a ten-society restricted-only qualification cohort on a separately sealed world reached 0.61--0.94 held-out compositional accuracy with universal communication dependence, but passed 0 of 10 complete preregistered gates: one model met every task-performance criterion, and all ten failed the ordinary-language preservation gate with 50--61 points of top-1 regression on general prompts. A subsequent restricted-versus-global performance contrast on development worlds motivated the paired, sealed, slot-selective-masking battery reported here.

\section{Preregistration, incidents, and compute lineage}\label{app:incidents}
All thresholds in Appendix~\ref{app:protocol} were frozen in writing before any training on the sealed instance. Training ran on rented consumer GPUs (RTX 5090 class for societies; A6000 class for staged comparators) admitted only after bit-exact golden-fingerprint reproduction. The incident ledger discloses: (i) one lane terminated silently by a provider host mid-training and rerun from scratch; (ii) a provider-account balance exhaustion that froze the fleet mid-campaign, recovered through a bit-exact resume protocol restoring trainable parameters, optimizer state, and all random-number-generator streams, with every affected final verified at internal step 20{,}000; (iii) one restricted final checkpoint (init 202, order 952) marooned on an unreachable provider instance, excluded from collision-stratified analysis and disclosed wherever relevant. Secondary staged-control runs also experienced checkpoint-filename reuse across successive resumes; their final lineage was reconstructed from the checkpoint-internal step field, and every reported final was verified at internal step 20{,}000. One transferred checkpoint copy failed checksum and was evaluated from the verified local copy. No checkpoint selection occurred at any point: all evaluations use final checkpoints only.

\section{Reproducibility and artifact release}\label{app:repro}
Training and evaluation run under deterministic settings (deterministic algorithms enforced; TF32 disabled; fixed gradient-accumulation schedule). The released package contains 19 of the 20 society final checkpoints and all three staged-comparator finals (adapter and projection weights); all evaluation, stratification, and audit JSON records; the audit and evaluation scripts; the preregistration and protocol-history documents; and the incident ledger. Code and records: \url{https://github.com/tokenosopher/populus-evidence-partitioning}; checkpoints: \url{https://huggingface.co/tokenosopher/populus-evidence-partitioning-checkpoints}. The remaining restricted checkpoint (initialization 202, order 952) is unavailable because its provider instance could not be recovered; its final evaluation records are included.

\begin{figure}[h]
\centering
\includegraphics[width=0.95\textwidth]{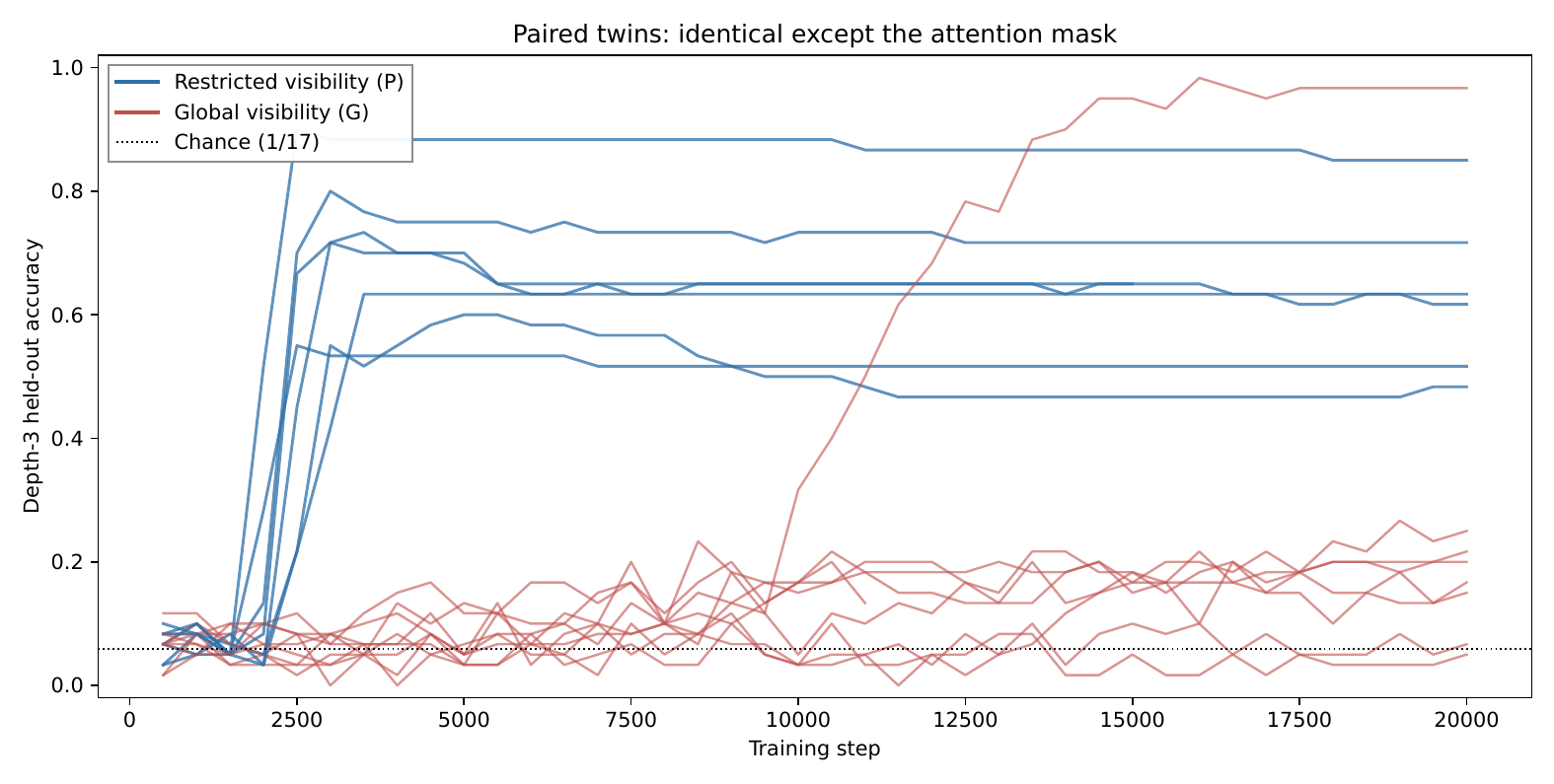}
\caption{\textbf{Training trajectories} (restricted and global arms), moved to the appendix per the main-figure policy: depth-three held-out probe accuracy over training. Curves are shown for 16 societies with complete per-500-step logs and 2 with truncated logs; 2 trajectories have no surviving probe log, lost to provider interruptions. Final-checkpoint evaluations are available for all 20 societies, while 19 final checkpoint files are currently recoverable.}
\label{fig:traj}
\end{figure}

\section{Cross-family portability of a value-indexed packet interface}\label{app:scout}
\emph{Added in v2.} After completing the main paired experiment, we prospectively froze a near-transfer scout asking whether the learned packet interface generalized to new surface language and primitive identities over the same 17-state carrier space. The scout protocol, criteria, and analysis plan were frozen before any evaluation, after an independent AI-assisted protocol review (not conventional journal peer review); we report every preregistered outcome. All numbers below use the same chance level ($1/17\approx0.0588$) and the same deterministic, machine-gated infrastructure as the main battery (Appendix~\ref{app:incidents}), with every admitted machine reproducing the golden fingerprints bit-exactly.

\paragraph{Design.} A second task family (family D) was generated and sealed before any checkpoint saw an episode: twelve fresh affine bijections over $\mathbb{Z}_{17}$ (forward steps, backward steps, reflections, and stretches; zero exact primitive-map overlap with the training family), rendered in a new templated language with fourteen phrasings per operator kind (bank hash \texttt{2bd0b301a5b079b1}). Family D deliberately preserves the 17-state cardinality, answer labels, relay anatomy, and affine algebra: it is a \emph{near-transfer} family by construction, changing what is said and which operations are meant while holding the carrier space fixed. One preregistration erratum is disclosed: the reflection primitive was initially implemented as $c-x$ against templates that describe reflection through a slot ($2c-x$); this was caught and corrected in pre-launch verification, and the banks were resealed, before any GPU evaluation. Evaluations cover the six restricted-visibility finals previously selected for causal auditing to span the observed performance range, plus the best-performing globally visible final (init 204, order 954) as a comparator.

\paragraph{Zero-shot and attributability.} The frozen base model with \emph{full} evidence visibility scores at chance on family D under raw and chat serialization (0.0588--0.0598), showing that it does not natively solve the target task zero-shot; this does not imply that the pretrained backbone is causally irrelevant to the learned representations or adaptation results below. All seven frozen societies likewise score at chance zero-shot (0.044--0.083 across depth-two and depth-three banks, including composition-map-novel strata), and severing all packets yields exactly 0.0588 in every case.

\paragraph{Carrier portability.} The central measurement transplants family-A-born packets into the family-D value/context layout while routing them through already learned family-A structural forwarders, balanced across interfaces and values, with counterfactual controls (donor packets for a value offset of $+8$). Across all six audited restricted-visibility checkpoints, family-A packet states remain perfectly transportable and causally steering in the family-D context (same-value preservation 1.000 and counterfactual following 1.000 in every model), whereas the sole globally visible comparator fails the same transplant test decisively (0.110 and 0.086, far below the preregistered 0.90 floor). This demonstrates near-transfer portability of the restricted models' value carrier. Crucially, communication necessity alone does not imply portability: the globally visible comparator required communication on its home task but did not expose an interchangeable cross-family value interface. Forcing known packets into the final interface shows the value-to-label mouth pathway is likewise fully functional in the new context (1.000 in family-D context, 1.000 in family-A control, 1.000 agreement, all seven models).

\paragraph{Failure localization.} Exhaustive single-primitive evaluation (12 operators $\times$ 14 phrasings $\times$ 3 slot positions $\times$ 17 start values $=8{,}568$ cases) scores 0.051--0.086 in all seven frozen societies: the frozen A-trained cells do not map family-D operation descriptions to the required local transformations. The exhaustive primitive test identifies local interpretation and execution of the new operator descriptions as at least one sufficient bottleneck in strict zero-shot transfer; forced-packet controls show that the carrier and final value-to-label pathway remain functional, but this scout did not independently isolate source-span parsing. A ridge probe on packet states under held-out programs finds no linearly decodable family-D running-value signal (0.033--0.067, within the observed permuted-baseline range of 0.017--0.100), consistent with the failure of the frozen local transducer to construct the target state; this diagnostic does not exclude a nonlinear or otherwise unusable representation.

\paragraph{Learnability gate.} A fresh restricted society (seeds committed before launch) trained on family D alone under the frozen protocol reaches held-out-phrasing accuracy 0.979, depth-two 0.922, and depth-three 0.917 at 20{,}000 updates, with all-cut exactly at chance and finite packets throughout, passing every preregistered learnability criterion; transfer results are therefore interpretable.

\paragraph{Frozen-interface transfer.} Six preregistered 5{,}000-update adaptation runs were conducted on two purposively selected restricted checkpoints (init 201/order 951 and init 203/order 953), with frozen per-checkpoint data streams identical across arms. A family-A packet interface can remain completely frozen while the shared LoRA learns the family-D local operations atop the frozen backbone: with reader, writer, mouth, and $\beta$ frozen and only the shared LoRA (plus a training-only auxiliary head) adapting, both checkpoints reached 0.66--0.69 depth-three accuracy within 5{,}000 updates, and cutting the frozen channel reduced performance to chance (preregistered criterion: $\ge0.40$ with all-cut $\le0.11$; passed in both checkpoints). The adaptation occurred on the same order of timescale as the fresh learnability run. In contrast, the preregistered initialization-benefit criterion \emph{failed} in both checkpoints: full adaptation from the trained interface versus adaptation with a freshly reinitialized reader/writer/mouth (atop the same A-trained LoRA) showed learning-curve-area differences of 0.001 and 0.042 and final differences of $-0.021$ and $+0.020$, far below the preregistered margins of 0.10. No preregistered initialization advantage was detected; this null concerns interface initialization conditional on the A-trained LoRA, across two purposively selected checkpoints and one data stream per checkpoint, and is not evidence that communication pretraining is generally valueless.

\paragraph{Subsequent work.} A standalone companion study (in preparation) extends this scout with leakage-controlled cross-model packet-transplant audits over all ordered society pairs, a source-span localization control, second-stream replications of the interface-inheritance comparison, and an interface-adaptation factorial on the globally visible checkpoint.

\paragraph{Verdict.} The scout reveals a layered form of transfer. The complete frozen society does not solve family D zero-shot. The exhaustive local-primitive test identifies inability to execute the new operation descriptions as a sufficient bottleneck, while source-span parsing was not independently isolated in this scout. Nevertheless, all six audited restricted-visibility checkpoints expose a family-A packet code that remains perfectly transportable and causally steering in the family-D context, while the sole globally visible comparator does not. Adapting only the shared LoRA, with reader, writer, mouth, and $\beta$ frozen, then yields substantial family-D composition in both tested checkpoints. However, the inherited interface provides no preregistered \emph{large} learning advantage over an interface reinitialized atop the same A-trained LoRA. The result therefore establishes near-transfer portability and frozen-interface sufficiency, not an algebra-independent universal protocol or a general benefit of protocol initialization. Concurrent work on decode-free bridges between heterogeneous frozen models \citep{yang2026xbridge} reports that bridges trained across several task domains generalize better than single-task bridges, and that continuous channels may require discrete anchoring to preserve entity identity; both findings inform the pretraining study this scout was designed to calibrate.

\paragraph{Replication and release.} One transfer arm was re-run end-to-end with identical seeds on a second independently admitted machine in a different datacenter: the full probe trajectory, final evaluations, batch-stream hash, and final checkpoint SHA-256 were identical across the two machines. The complete scout --- family-D generation and preflight, all portability and localization evaluations, the learnability run, and all six transfer arms --- consumed under \$10 of rented consumer-GPU time. All scout evaluation records, the sealed family-D module, the frozen plan, and the adaptation checkpoints are released with the artifact package (Appendix~\ref{app:repro}).

\end{document}